\documentclass[11pt,a4paper]{article}
\usepackage[hyperref]{emnlp2020}
\usepackage{times}
\usepackage{latexsym}

\usepackage{microtype}

\usepackage{enumitem}
\usepackage{graphicx}
\usepackage{float}

\usepackage{amssymb}
\usepackage{multirow}

\usepackage{booktabs}
\usepackage{multirow}

\usepackage{pbox}
\usepackage{amsmath}

\DeclareMathOperator*{\argmin}{arg\,min}
\usepackage{url}

\newcommand{\fontmed}{\fontsize{9pt}{9pt}\selectfont}

\usepackage[normalem]{ulem}
\useunder{\uline}{\ul}{}

\usepackage{siunitx}

\aclfinalcopy 

\title{Self-Supervised Knowledge Triplet Learning for Zero-shot \\Question Answering}

\author{Pratyay Banerjee  \and Chitta Baral 
\\ Department of Computer Science, Arizona State University
\\ \texttt{pbanerj6,chitta}@asu.edu
}

\date{}

\begin{document}
\maketitle
\begin{abstract}
The aim of all Question Answering (QA) systems is to be able to generalize to unseen questions. Current supervised methods are reliant on expensive data annotation. Moreover, such annotations can introduce unintended annotator bias which makes systems focus more on the bias than the actual task. In this work, we propose Knowledge Triplet Learning (KTL), a self-supervised task over knowledge graphs. We propose heuristics to create synthetic graphs for commonsense and scientific knowledge.  We propose methods of how to use KTL to perform zero-shot QA and our experiments show considerable improvements over large pre-trained transformer models. 
\end{abstract}

\section{Introduction}

The ability to understand natural language and answer questions is one of the core focuses in the field of natural language processing. To measure and study the different aspects of question answering, several datasets are developed, such as SQuaD \cite{rajpurkar2018know}, HotpotQA \cite{yang2018hotpotqa}, and Natural Questions \cite{kwiatkowski2019natural} which require systems to perform extractive question answering. On the other hand, datasets such as SocialIQA \cite{sap2019socialiqa}, CommonsenseQA \cite{talmor2018commonsenseqa}, Swag \cite{zellers2018swag} and Winogrande \cite{sakaguchi2019winogrande} require systems to choose the correct answer from a given set. These multiple-choice question answering datasets are very challenging, but recent large pre-trained language models such as BERT \cite{devlin2018bert}, XLNET \cite{yang2019xlnet} and RoBERTa \cite{liu2019roberta} have shown very strong performance on them. Moreover, as shown in Winogrande \cite{sakaguchi2019winogrande}, acquiring unbiased labels requires a ``carefully designed crowdsourcing procedure'', which adds to the cost of data annotation. This is also quantified in other natural language tasks such as Natural Language Inference \cite{gururangan2018annotation}  and Argument Reasoning Comprehension \cite{niven2019probing}, where such annotation artifacts lead to ``Clever Hans Effect'' in the models \cite{kaushik2018much,poliak2018hypothesis}. One way to resolve this is to design and create datasets in a clever way, such as in Winogrande \cite{sakaguchi2019winogrande}, another way is to ignore the data annotations and to build systems to perform unsupervised question answering \cite{teney2016zero,lewis-etal-2019-unsupervised}. In this paper, we focus on building unsupervised zero-shot multiple-choice QA systems. 

\begin{figure}[t]
\includegraphics[width=0.95\linewidth]{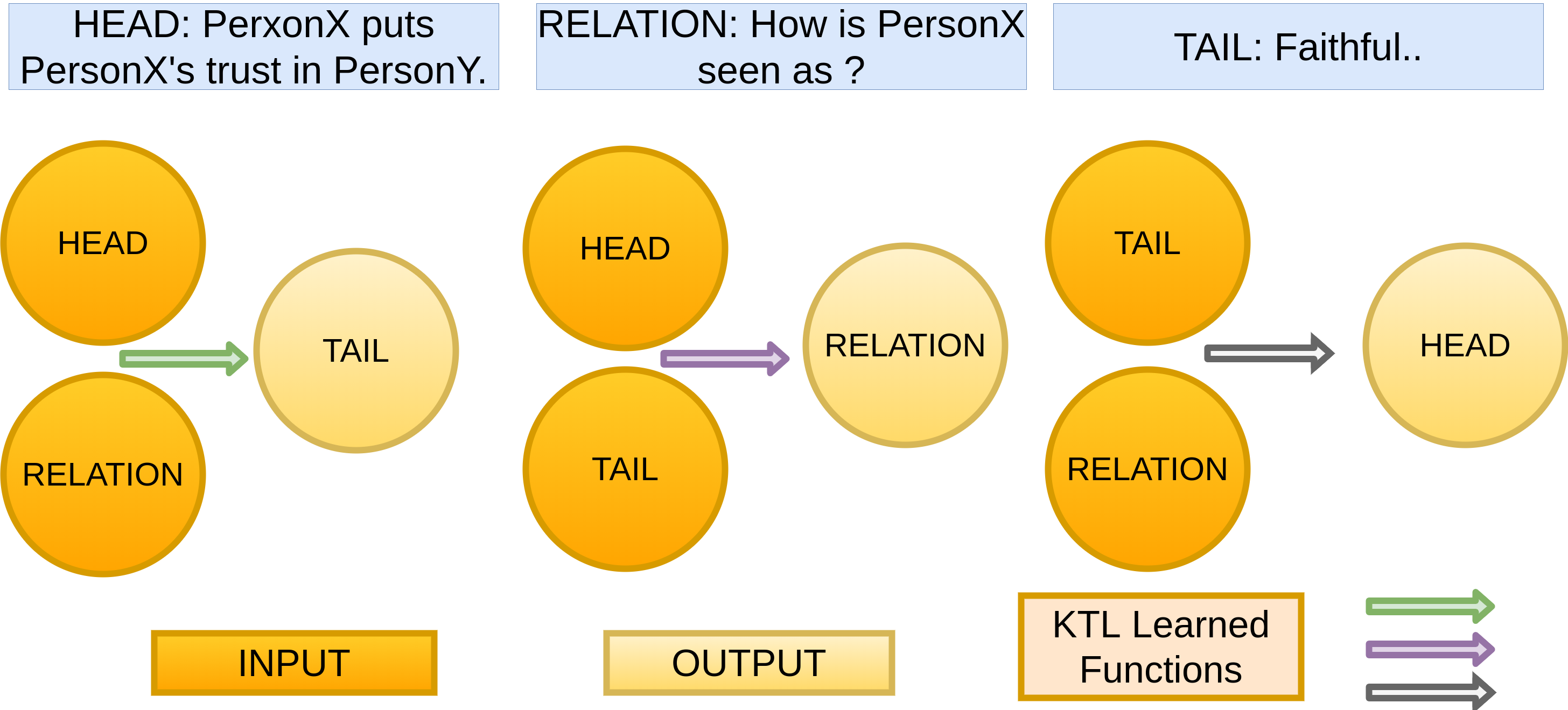}
\caption{Knowledge Triplet Learning Framework, where given a triple $(h,r,t)$ we learn to generate one of the inputs given the other two. }
\label{fig:1}
\end{figure}

Recent work \cite{fabbri2020template,lewis-etal-2019-unsupervised} try to generate a synthetic dataset using a text corpus such as Wikipedia, to solve extractive QA. Other works \cite{bosselut2019dynamic,shwartz2020unsupervised} use large pre-trained generative language models such as GPT-2 \cite{radford2019language} to generate knowledge, questions, and answers and compare against the given answer choices.

In this work, we utilize the information present in Knowledge Graphs such as ATOMIC \cite{Sap2019ATOMICAA}. We define a new task of Knowledge Triplet Learning (KTL) over these knowledge graphs. For tasks which do not have appropriate knowledge graphs, we propose heuristics to create synthetic knowledge graphs. Knowledge Triplet Learning is like Knowledge Representation Learning and Knowledge Graph Completion but not limited to it. Knowledge Representation Learning \cite{lin2018knowledge} learns the low-dimensional projected and distributed representations of entities and relations defined in a knowledge graph. Knowledge Graph Completion \cite{ji2020survey} aims to identify new relations and entities to expand an incomplete input knowledge graph. 

In KTL, as shown in Figure \ref{fig:1}, we define a triplet $(h, r, t)$, and given any two as input we learn to generate the third. This tri-directional reasoning forces the system to learn all the possible relations between the three inputs. 
We map the question answering task to KTL, by mapping the \textit{context}, \textit{question} and \textit{answer} to $(h,r,t)$ respectively. 
We define two different ways to perform self-supervised KTL. This task can be designed as a representation generation task or a masked language modeling task. We compare both the strategies in this work. We show how to use models trained on this task to perform zero-shot question answering without any additional supervision. We also show how models pre-trained on this task perform considerably well compared to strong pre-trained language models on few-shot learning. We evaluate our approach on the three commonsense and three science multiple-choice QA datasets. 

The contributions of this paper are summarized as follows:
\begin{itemize}[noitemsep]
    \item We define the Knowledge Triplet Learning over Knowledge Graph and show how to use it for zero-shot question answering. 
    \item We compare two strategies for the above task.
    \item We propose heuristics to create synthetic knowledge graphs.
    \item We perform extensive experiments of our framework on three commonsense and three science question answering datasets.
    \item We achieve state-of-the-art results for zero-shot and propose a strong baseline for the few-shot question answering task. 
\end{itemize}

\section{Knowledge Triplet Learning}
We define the task of Knowledge Triplet Learning (KTL) in this section. We define $G = (V,E)$ as a Knowledge Graph, where $V$ is the set of vertices, $E$ is the set of edges. $V$ consists of entities which can be phrases or named-entities depending on the given input Knowledge Graph. Let $S$ be a set of fact triples,  $S \subseteq V \times E \times V$  with the format $(h,r,t)$, where $h$ and $t$ belong to set of vertices $V$ and $r$ belongs to set of edges. The $h$ and $t$ indicates the head and tail entities, whereas $r$ indicates the relation between these entities. 

For example, from the ATOMIC knowledge graph, (\textit{PersonX puts PersonX's trust in PersonY}, \textbf{How is PersonX seen as?}, \textit{faithful}) is one such triple. Here the head is \textit{PersonX puts PersonX's trust in PersonY}, relation is \textbf{How is PersonX seen as?} and the tail is \textit{faithful}. Do note $V$ does not contain homogenous entities, i.e, both \textit{faithful} and \textit{PersonX puts PersonX's trust in PersonY} are in $V$.

We define the task of KTL as follows:
Given input a triple $(h,r,t)$, we learn the following three functions.
\begin{equation}
    \begin{split}
     f_t(h,r) \Rightarrow t, \;\;
     f_h(r,t) \Rightarrow h, \;\;
     f_r(h,t) \Rightarrow r 
    \end{split}
\label{eq:gen}
\end{equation}
That is, each function learns to generate one component of the triple given the other two. The intuition behind learning these three functions is as follows. Let us take the above example: (\textit{PersonX puts PersonX's trust in PersonY}, \textbf{How is PersonX seen as?}, \textit{faithful}). The first function $f_t(h,r)$ learns to generate the answer $t$ given the context and the question. The second function $f_h(r,t)$ learns to generate one context where the question and the answer may be valid. The final function $f_r(h,t)$ is a Jeopardy-style generating the question which connects the context and the answer.

In Multiple-choice QA, given the context, two choices may be true for two different questions. Similarly, given the question, two answer choices may be true for two different contexts. For example, given the context: \textit{PersonX puts PersonX's trust in PersonY}, the answers \textit{PersonX is considered trustworthy by others} and \textit{PersonX is polite} are true for two different questions \textbf{How does this affect others?} and \textbf{How is PersonX seen as?}. Learning these three functions enables us to score these relations between the context, question, and answers.

\subsection{Using KTL to perform QA}
After learning this function in a self-supervised way, we can use them to perform question answering.
Given a triple $(h,r,t)$, we define the following scoring function:
\begin{equation}
\fontmed
\begin{split}
    & D_t = D(t,f_t(h,r)), \;\; D_h = D(h,f_h(r,t)), \\
    & D_r = D(r,f_r(h,t)) \\
    & score(h,r,t) = D_t * D_h * D_r \\
\end{split}
\label{eq:dist}
\end{equation}
where $h$ is the context, $r$ is the question and $t$ is one of the answer options. $D$ is a distance function which measures the distance between the generated output and the ground-truth. The distance function varies depending on the instantiation of the framework, which we will study in the following sections. The final answer is selected as:
\begin{equation}
    ans = \argmin_t(score(h,r,t))
\label{eq:score}
\end{equation}
As the scores are the distance from the ground-truth we select the choice that has the minimum score.

We define the different ways we can implement this framework in the following sections.

\subsection{Knowledge Representation Learning}
In this implementation, we use Knowledge representation learning to learn equation \eqref{eq:gen}. In contrast to triplet classification and graph completion, where systems try to learn a score function $f_r(h,t)$, i.e, is the fact triple $(h,r,t)$ true or false; in this method we learn to generate the inputs vector representations, i.e, $f_r(h,t) \Rightarrow r$. We can view equation \ref{eq:gen} as generator functions, which given the two input encodings learns to generate a vector representation of the third. 
As our triples $(h,r,t)$ can have a many to many relations between each pair, we first project the two inputs from input encoding space to a different space similar to the work of TransD \cite{ji2015knowledge}. We use a Transformer encoder $Enc$ to encode our triples to the encoding space. We learn two projection functions, $M_{i1}$ and $M_{i2}$ to project the two inputs, and a third projection function $M_o$ to project the entity to be generated. We combine the two projected inputs using a function $C$. These functions can be implemented using feedforward networks. 
\begin{equation*}
\fontmed
    \begin{split}
        & I_{e1} = Enc(I_1),  
         I_{e2} = Enc(I_2),  
         O_e = Enc(O) \\
        & I_{e1} = M_{i1}(I_{e1}), 
         I_{e2} = M_{i2}(I_{e2}), 
         O_{p} = M_{o}(O_e)\\
        & \Hat{O} = C(I_{e1},I_{e2}) \\
        & loss = LossF(\Hat{O},O_{p})
    \end{split}
\end{equation*}
where $I_i$ is the input, $\hat{O}$ is the generated output vector and $O_p$ is the projected vector. $M$ and $C$ functions are learned using fully connected networks. In our implementation, we use RoBERTa as the $Enc$ transformer, with the output representation of the $[cls]$ token as the phrase representation.

We train this model using two types of loss functions, L2Loss where we try to minimize the L2 norm between the generated and the projected ground-truth, and Noise Contrastive Estimation \cite{gutmann2010noise} where along with the ground-truth we have $k$ noise-samples. These noise samples are selected from other $(h,r,t)$ triples such that the target output is not another true fact triple, i.e, $(h,r,t_{noise})$ is false.
The NCELoss is defined as:
\begin{equation*}
\fontmed
\begin{split}
    & NCELoss(\Hat{O},O_{p},[N_0...N_k]) =  \\
    & - \log \frac{\exp{sim(\Hat{O},O_{p})}}{\exp{sim(\Hat{O},O_{p})} + \sum_{k \in N}{\exp{(sim(\Hat{O},N_k)}}}
\end{split}
\end{equation*}
where $N_k$ are the projected noise samples, $sim$ is the similarity function which can be the L2 norm or Cosine similarity, $\hat{O}$ is the generated output vector and $O_p$ is the projected vector.

The $D$ distance function \eqref{eq:dist} for such a model is defined by the distance function used in the loss function. For L2Loss, it is the L2 norm, and in the case of NCELoss, we use $1-sim$ function.  

\subsection{Span Masked Language Modeling}
In Span Masked Language Modeling (SMLM), we model the equation \ref{eq:gen} as a masked language modeling task. We tokenize and concatenate the triple $(h,r,t)$ with a separator token between them, i.e, $[cls][h][sep][r][sep][t][sep]$. For the function $f_r(h,t) \Rightarrow r$, we mask all the tokens present in $r$, i.e, $[cls][h][sep][mask][sep][t][sep]$. We feed these tokens to a Transformer encoder $Enc$ and use a feed forward network to unmask the sequence of tokens. Similarly, we mask $h$ to learn $f_h$ and $t$ to learn $f_t$ 

We train the same Transformer encoder to perform all the three functions. We use the cross-entropy loss to train the model:
\begin{equation*}
\fontmed
\begin{split}
     &   CELoss(h,r,mask(t),t) = \\
     & - \frac{1}{n}\sum_{i=1}^{n}log_2P_{MLM}(t_i|h,r,t_{1..t_i..tn}) 
\end{split}
\end{equation*}
where $P_{MLM}$ is the masked language modeling probability of the token $t_i$, given the unmasked tokens $h$ and $r$ and other masked tokens in $t$. Do note we do not do progressive unmasking, i.e, all the masked tokens are jointly predicted.

The $D$ distance function \eqref{eq:dist} for this model is same as the loss function defined above. 

\section{Synthetic Graph Construction}
In this section we describe our method to create a synthetic knowledge graph from a text corpus containing sentences. Not all types of knowledge are present in a structured knowledge graph, such as ATOMIC, which might be useful to answer questions. For example, the questions in QASC dataset \cite{khot2019qasc} require knowledge about scientific concepts, such as, ``Clouds regulate the global engine of atmosphere and ocean.". The QASC dataset contains a textual knowledge corpus containing science facts. Similarly, the Open Mind Commonsense (OMCS) knowledge corpus contains knowledge about different commonsense facts, such as, ``You are likely to find a jellyfish in a book". Another kind of knowledge about social interactions and story progression is present in several story understanding datasets, such as RoCStories and the Story Cloze Test \cite{mostafazadeh2016corpus}. To be able to perform question answering using these knowledge and KTL, we create the following two graphs: Common Concept Graph and the Directed Story Graph. 

\paragraph{Common Concept Graph}
In order to create the Common Concept Graph, we extract noun-chunks and verb-chunks from each of the sentences using Spacy Part-of-Speech tagger \cite{spacy2}. We assign all the extracted chunks as the vertices of the graph, and the sentences as the edges of the graph. To generate training samples for KTL, we assign triples $(h,R,t)$ as  $(e_1,e_2,v_i)$ where $v_i$ is the common concept present in both the sentences $e_1$ and $e_2$. 
For example, in the sentence \textit{Clouds regulate the global engine of atmosphere and ocean.}, the extracted concepts are \textit{clouds}, \textit{global engine}, \textit{atmosphere}, \textit{ocean} and \textit{regulate}. The triplet assignment will be, [\textit{Warm moist air from the Pacific Ocean brings fog and low stratus clouds to the maritime zone.}, \textit{Clouds regulate the global engine of atmosphere and ocean.}, \textbf{clouds}]. We create two such synthetic graphs using the QASC science corpus and the OMCS concept corpus. Our hypothesis is this graph and the KTL framework will allow the model to understand the concepts common in two facts, which in turn allows question answering.

\paragraph{Directed Story Graph}
This graph is created using short stories from the RoCStories and Story Cloze Test datasets. This graph is different from the above graph as this graph has a directional property and each individual story graph is disconnected. To create this graph, we take each short story with $k$ sentences, $[s_1,s_2,s_3..,s_k]$ and create a directed graph such that, all sentences are vertices and each sentence is connected with a directed edge only to sentences that occur after it. For example, $s_1$ is connected to $s_2$ with a directed edge but not vice versa. We generate triples $(h,R,t)$ by sampling vertices $(s_i,s_j,s_k)$ such that there is a directed path between the sentences $s_i$ and $s_k$ through $s_j$. This captures a smaller story where the head is an event that occurs before the relation and the tail. This graph is designed for story understanding and abductive reasoning using KTL framework.

\paragraph{Random Sampling}
There are around 17M sentences in the QASC text corpus, similarly there are 640K sentences in the OMCS text corpus. Our synthetic triple generation leads to a significantly large set of triples that is in order of $10^{12}$ and more. To restrict the train dataset size for our KTL framework, we randomly sample triples and limit the train dataset size to be at max 1M samples, we refer to this as Random Sampling.

\paragraph{Curriculum Filtering} Here we extract the noun and verb chunks from the context, question and answer options present in the question answering datasets. We filter triples from the generated dataset and keep only those triples where at least one of the entities is present in the extracted noun and verb chunks set. This filtering is analogous to a real-life human examination setting where a teacher provides the set of concepts upon which questions would be asked, and the students can learn the concepts. We perform the sampling and filtering only on the huge Common Concept Graphs generated from QASC and OMCS corpus. 

\begin{table*}[t]
\centering
\small
\begin{tabular}{@{}lllllllll@{}}
\toprule
 &  & \textbf{ARC-Easy} & \textbf{ARC-Chall} & \textbf{QASC} & \textbf{OpenBookQA} & \textbf{CommonsenseQA} & \textbf{aNLI} & \textbf{SocialIQA} \\ \midrule
 & Train Size & 2251 & 1119 & 8134 & 4957 & 9741 & 169654 & 33410 \\
 & Val Size & 570 & 299 & 926 & 500 & 1221 & 1532 & 1954 \\ 
 & Test Size & 2377 & 1172 & 920 & 500 & 1140 & - & - \\ 
 & C Length & - & - & - & - & - & 9 & 15 \\ 
 & Q Length & 19.4 & 22.3 & 13 & 12 & 14 & 9 & 6 \\ 
 & A length & 3.7 & 4.9 & 1.5 & 3 & 1.5 & 9 & 3 \\ 
 & \# of Option & 4 & 4 & 8 & 4 & 5 & 2 & 3 \\ 
 & KTL Graph & QASC-CCG & QASC-CCG & QASC-CCG & QASC-CCG & OMCS-CCG & DSG & ATOMIC \\ \bottomrule
\end{tabular}%
\caption{Dataset Statistics for the seven QA tasks. Context is not present in five of the tasks. The KTL Graph refers to the graph over which we learn. CCG is the Common Concept Graph. DSG is the Directed Story Graph. C,Q,A is the average number of words in the context, question and answer. aNLI and SocialIQA Test set size is hidden. }
\label{tab:dstats}
\end{table*}

\begin{table}[t]
\centering
\resizebox{\linewidth}{!}{%
\begin{tabular}{@{}lllll@{}}
\toprule
 & \textbf{ATOMIC} & \textbf{QASC-CCG} & \textbf{OMCS-CCG} & \textbf{DSG} \\ \midrule
Train Size & 893393 & 1662308 & 914442 & 1019030 \\
Val Size & 10000 & 10000 & 10000 & 10000 \\
H Length & 11.2 & 10.5 & 9.6 & 10.3 \\
R Length & 6.5 & 10.3 & 9.4 & 10.2 \\
T Length & 2 & 1.5 & 2 & 10.4 \\ \bottomrule
\end{tabular}%
}
\caption{Dataset Statistics for the generated Triples. For QASC and OMCS it is after Curriculum Filtering. H,R,T length refers to average number of words. For CCG, we show for the $[e_i,e_j,v]$ configuration.}
\label{tab:dstatsgraphs}
\end{table}

\section{Datasets}
We evaluate our framework on the following six datasets: SocialIQA \cite{sap2019socialiqa}, aNLI \cite{bhagavatula2019abductive}, CommonsenseQA \cite{talmor2018commonsenseqa}, QASC \cite{khot2019qasc}, OpenBookQA \cite{mihaylov2018can} and ARC \cite{clark2018think}. SocialIQA, aNLI and CommonsenseQA require commonsense reasoning and external knowledge to be able to answer the questions. Similarly, QASC, OpenBookQA, and ARC require scientific knowledge. Table \ref{tab:dstats} shows the dataset statistics and the corresponding knowledge graph used to train our KTL model. Table \ref{tab:dstatsgraphs} shows the statistics for the triples extracted from the graphs. From the two tables we can observe our KTL triples have different number of words when compared to the target question answering tasks. Especially, where the context is significantly larger and human annotated as in SocialIQA, increasing the challenge for unsupervised learning. 

\subsection{Question to Hypothesis Conversion and Context Creation}
We can observe the triples in our synthetic graphs, QASC-CCG and OMCS-CCG contain factual statements, and our target question answering datasets have questions that contain \emph{wh} words or fill-in-the-blanks. We translate each question to a hypothesis using the question and each answer option. To create hypothesis statements for questions containing \emph{wh} words, we use a rule-based model \cite{Demszky2018TransformingQA}. For fill-in-the-blank and cloze style questions, we replace the blank or concat the question and the answer option.

For questions that do not have a context, such as in QASC or CommonsenseQA, we retrieve top five sentences using the question and answer options as query and perform retrieval from respective source knowledge sentence corpus. For each retrieved context, we evaluate the answer option score using equation \ref{eq:dist} and take the mean score. 

\begin{table*}[t]
\centering
\small
\begin{tabular}{@{}llllllll@{}}
\toprule
 \textbf{Models} & \textbf{ARC-E $\uparrow$} & \textbf{ARC-C} $\uparrow$ & \textbf{OBQA} $\uparrow$ & \textbf{QASC} $\uparrow$ & \textbf{ComQA} $\uparrow$ & \textbf{ \; aNLI} $\uparrow$ & \textbf{SocIQA} $\uparrow$ \\ \midrule
Random     & 25.0  25.0  25.0 & 25.0  25.0  25.0 & 25.0 25.0 25.0 & 12.5 12.5 & 20.0 20.0 & 50.0 51.0 & 33.3 33.3 \\
GPT-2 L      & 30.5  29.1  29.4 & 23.5  25.1  25.0 & 32.0 26.6 27.8 & 12.3 13.2 & 36.4 37.2 & 50.8 51.3 & 41.2 40.8 \\ 
RoB-MLM    & 29.8  29.6  29.0 & 24.8  25.0  25.0 & 24.8 24.4 25.0 & 12.8 17.6 & 23.6 24.8 & 51.6 52.2 & 35.6 34.5 \\ 
RoB-FMLM   & 31.0  31.2  30.6 & 24.6  22.1  23.8 & 23.4 24.2 23.8 & 14.2 19.7 & 23.2 26.1 & 51.2 51.4 & 36.9 36.1 \\ 
IR         & 29.4  30.4  30.2 & 18.4  20.3  21.2 & 31.4 29.4 28.8 & 18.6 19.4 & 24.6 24.4 & 53.4 54.8 & 35.8 36.0 \\ \midrule
KRL-L2     & 28.8  29.6  29.8 & 26.7  26.8  25.6 & 29.6 28.8 29.2 & 20.4 20.8 & 31.4 30.6 & 57.6 57.4 & 43.2 43.8 \\ 
KRL-NCE-L2 & 32.4  31.8  30.6 & 27.2  27.5  26.8 & 33.2 31.6 32.8 & 22.6 23.1 & 33.4 33.8 & 59.3 60.5 & 46.4 46.2 \\ 
KRL-NCE-Cos&  \underline{32.8} \underline{32.0}  \underline{31.8} &
              \underline{27.4} \underline{27.9}  \underline{27.8} &
              \textbf{35.6 34.8 34.4}                             &
              \underline{23.2} \underline{24.4} &
              \underline{36.8} \underline{37.1} &
              \underline{60.4} \underline{60.2} &
              \underline{46.6} \underline{46.4} \\ 
SMLM &        \textbf{33.2  33.4 33.0 }                    &
              \textbf{27.8  28.4 28.4 }                    &
              \underline{34.4} \underline{34.6}  \underline{33.8}     & 
              \textbf{26.6 27.2}                      &
              \textbf{38.2 38.8}                      &
              \textbf{64.7 65.3}                      &
              \textbf{48.7 48.5} \\ 
\midrule
Self-Talk & \hphantom{1234} N/A & \hphantom{1234} N/A   & \hphantom{1234} N/A  & \hphantom{12} N/A & \hphantom{12}  32.4 & \hphantom{12} N/A  &  \hphantom{12} 46.2 \\ 
BIDAF Sup. & \hphantom{12} 50.1  49.8 & \hphantom{12}  20.6  21.2 & \hphantom{12}  49.2 48.8 & \hphantom{12} 31.8 & \hphantom{12} 32.0 & \hphantom{12} 67.8 & \hphantom{12}  51.2 \\
RoBerta Sup. & \hphantom{1234} 85.0 & \hphantom{1234} 67.2 & \hphantom{1234} 72.0 & \hphantom{12} 61.8 &\hphantom{12} 72.1 & \hphantom{12} 83.2 & \hphantom{12} 76.9\\
\bottomrule
\end{tabular}%
\caption{Results for the Unsupervised QA task. Mean accuracy on Train, Dev and Test is reported. For Self-Talk and BIDAF Sup. we report the Dev and Test splits, for Roberta Sup. we report Test split. Test is reported if labels are present. \textbf{Best} scores, \underline{Second Best}.}
\label{tab:unsupcomp}
\end{table*}


\section{Experiments}
\subsection{Baselines}
We compare our models to the following baselines.
\begin{enumerate}[nosep, noitemsep]
    \item \textbf{GPT-2 Large} with language modeling cross-entropy loss as the scoring function. We concatenate the context and question and find the cross-entropy loss for each of the answer choices and choose the answer which has the minimum loss. 
    \item \textbf{Pre-trained RoBerta-large} used as is, without any finetuning or further pre-training, with scoring same as our defined SMLM model. We refer to it as Rob-MLM. 
    \item \textbf{RoBerta-large} model further fine-tuned using the original Masked Language Modeling task over our concatenated fact triples $(h,r,t)$, with scoring same as SMLM.  We refer to it as Rob-FMLM.
    \item \textbf{IR Solver} described in ARC \cite{clark2016combining}, which sends the context, question and answer option as a query to Elasticsearch. The top retrieved sentence which has a non-stop-word overlap with both the question and the answer is used as a representative, and its corresponding IR ranking score is used as confidence for the answer. The option with the highest score is chosen as the answer. 
\end{enumerate}

\subsection{KTL Training}
We train the Knowledge Representation Learning (KRL) model using both L2Loss and NCELoss. For NCELoss we also train it with both L2 norm and Cosine similarity. Both the KRL model (365M) and SMLM model (358M) uses RoBERTa-large (355M) as the encoder. We train the model for three epochs with the following hyper-parameters: batch sizes [512,1024] for SMLM and [32,64] for KRL; learning rate in range: [1e-5,5e-5]; warm-up steps in range [0,0.1]; in 4 Nvidia V100s 16GB. We use the transformers package \cite{Wolf2019HuggingFacesTS}. All triplets from the training graphs are positive samples. We learn using these triplets. For NCE, we choose $k$ equal to ten, i.e., ten negative samples. We perform three hyper-parameter trials using ten percent of the training data for each model, and train models with three different seeds. For each of our experiments we report the mean accuracy of the three random seed runs, and report the standard deviation if space permits. Code is available \href{https://www.github.com/pratyay-banerjee}{here}.

\section{Results and Discussion}

\subsection{Unsupervised Question Answering}
Table \ref{tab:unsupcomp} compares our different KTL methods with our four baselines for the six question answering datasets on the zero-shot question answering task. We use Hypothesis Conversion, Curriculum Filtering and Context Creation for ARC, QASC, OBQA and CommonsenseQA for both, the baselines as well as our models. We compare the models on the Train, Dev and Test split if labels are available, to better capture the statistical significance. 

We can observe our KTL trained models perform statistically significantly better than the baselines. When comparing the different KRL models, the NCELoss with Cosine similarity performs the best. This might be due to the additional supervision provided by the negative samples as the L2Loss model only tries to minimize the distance between the generated and the target projections. When comparing different KTL instantiations, we can see the SMLM model performs the best overall. SMLM and KRL differ in their core approaches. Our hypothesis is the multi-layered attention in a transformer encoder enables the SMLM model to better distinguish between a true and false statement. In KRL we are learning from both positive and negative samples, but the model still under-performs. On analysis, we observe the random negative samples may make the training task biased for KRL. Our future work would be to utilize alternative negative sampling techniques, such as selecting samples that are closer in contextual vector space.  

The improvements on ARC-Challenge task is considerably less. It is observed that the fact corpus for QASC, although contains a huge number of science facts, does not contain sufficient knowledge to answer ARC questions. There is a substantial improvement in SocialIQA, aNLI, QASC and CommonsenseQA as the respective KTL knowledge corpus contains sufficient knowledge to answer the questions. It is interesting to note that for QASC, we can reduce the problem from an eight-way to a four-way classification, as our top-4 accuracy on QASC is above 92\%. Our unsupervised model outperforms previous approaches such as Self-Talk \cite{shwartz2020unsupervised}. It approaches prior supervised approaches like BIDAF \cite{Seo2017BidirectionalAF}, and even surpasses it on two tasks.

\begin{table}[t]
\centering
\resizebox{\linewidth}{!}{%
\begin{tabular}{@{}llllll@{}}
\toprule
\textbf{Model} & \textbf{QASC} $\uparrow$  & \textbf{OBQA}  $\uparrow$  & \textbf{aNLI} $\uparrow$  & \textbf{ComQA} $\uparrow$  & \textbf{SocIQA} $\uparrow$ \\ \midrule
RoBerta & 44.5 $\pm$ 1.2 & 47.8 $\pm$ 1.4 & 68.8 $\pm$ 1.3 & 46.4 $\pm$ 1.5 & 44.4 $\pm$ 1.2 \\
RoB-MLM & 43.6 $\pm$ 0.6 & 49.4 $\pm$ 0.8 & 67.1 $\pm$ 0.8 & 43.2 $\pm$ 0.8 & 46.8 $\pm$ 0.6 \\
KRL-NCE-Cos & 48.2 $\pm$ 0.9 & 51.2 $\pm$ 0.6 & 73.4 $\pm$ 0.9 & 49.5 $\pm$ 1.1 & 58.6 $\pm$ 0.8 \\
SMLM & \textbf{49.8} $\pm$ 0.6 & \textbf{55.8} $\pm$ 0.6 & \textbf{76.8} $\pm$ 0.6 & \textbf{51.2} $\pm$0.7 & \textbf{69.1} $\pm$ 0.4 \\
RoBerta-Sup & 59.40 & 71.0 & 84.3 & 71.4 & 76.6 \\ \bottomrule
\end{tabular}%
}
\caption{Accuracy comparison of the KTL pre-trained RoBerta encoder when used for Few-shot learning Question Answering task on the Validation split. }
\label{tab:fewshot}
\end{table}

\begin{figure*}[t]
\includegraphics[width=\textwidth,height=3cm]{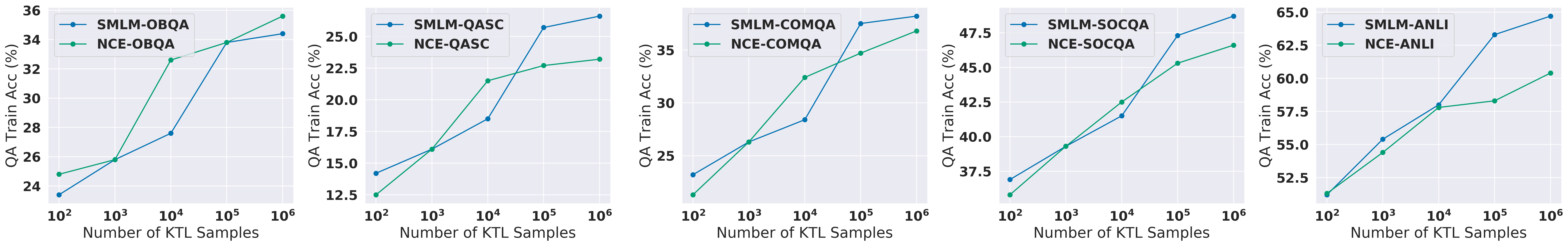}
\caption{Effect of Increasing KTL training samples on the target zero-shot question answering  Train split accuracy.}
\label{fig:trainingcurve}
\end{figure*}

\subsection{Few-Shot Question Answering}
Table \ref{tab:fewshot} compares our KTL pre-trained transformer encoder in the few-shot question answering task. We fine-tune the encoder with a simple feedforward network for a $n$-way classification task, the standard question answering approach using RoBerta with $n$ being the number of answer options, while training with only 8\% of the training data. We train on three randomly sampled splits of training data and report the mean. We can observe our KTL pre-trained encoders perform significantly better than the baselines, and approaches the fully supervised model, with only 7.5\% percent behind the fully supervised model on SocialIQA. We also observe our pre-trained models have a lower deviation.

\begin{table}[t]
\centering
\resizebox{\linewidth}{!}{%
\begin{tabular}{@{}llllll@{}}
\toprule
\textbf{Model} & \textbf{QASC} $\uparrow$ & \textbf{OBQA} $\uparrow$ & \textbf{ComQA} $\uparrow$ & \textbf{aNLI} $\uparrow$ & \textbf{SocIQA} $\uparrow$ \\ \midrule
SMLM - A      & 23.4 $\pm$ 0.6 & 28.6 $\pm$ 0.7 & 33.6 $\pm$ 0.5 & 64.8 $\pm$ 0.9 & 46.2 $\pm$ 0.7 \\
SMLM - Q      & 26.7 $\pm$ 0.8 & 33.8 $\pm$ 0.7 & 34.4 $\pm$ 0.8 & 65.1 $\pm$ 0.7 & 37.8 $\pm$ 0.5 \\
SMLM - C      & 22.8 $\pm$ 1.1 & 29.8 $\pm$ 1.3 & 31.9 $\pm$ 0.9 & 64.9 $\pm$ 0.8 & 47.1 $\pm$ 0.8 \\
SMLM - A*Q*C  & 27.2 $\pm$ 0.6 & 34.6 $\pm$ 0.8 & 38.8 $\pm$ 0.6 & 65.3 $\pm$ 0.7 & 48.5 $\pm$ 0.6 \\ \bottomrule
\end{tabular}%
}
\caption{Accuracy comparison of using only Answer (A), Question (Q) and Context (C) distance scores. }
\label{tab:absdistscores}
\end{table}

\subsection{Ablation studies and Analysis}
\paragraph{Effect of Context, Question, Answer Distance}
In Table \ref{tab:absdistscores} we compare the effect of the three different distance scores. It is interesting to observe, in OpenBookQA, QASC and CommonsenseQA, the three datasets which don't provide a context, the model is more perplexed to predict the question when given a wrong answer option, leading to higher accuracy for only Question distance score. On the other hand, in aNLI all three distance scores have nearly equal performance. In SocialIQA, the question has the least accuracy, whereas the model is more perplexed when predicting the context given a wrong answer option. This confirms our hypothesis that given a task predicting context and question can contain more information than discriminating between options alone. 

\paragraph{Effect of Hypothesis Conversion, Curriculum Filtering and Context Retrieval}
In Table \ref{tab:abshypcf} we observe the effect of hypothesis conversion, curriculum filtering and our context creation. Converting the question to a hypothesis provides a slight improvement, but a major improvement is observed when we filter our KTL training samples and keep only those concepts which are present in the target question answering task, compared to when the KTL model is trained with a random sample of 1M.  Curriculum filtering is impactful because there are a lot of concepts which are present in our source knowledge corpus, and the random sampled training corpus only contains 50\% of the target question answering task concepts on an average.  
Another key thing to note in Table \ref{tab:abshypcf} is our KTL models can strongly perform like supervised models, when the gold knowledge context is provided, which are available in QASC and OpenBookQA. This indicates a better retrieval system for context creation can further improve our models.   

\begin{table}[t]
\centering
\small
\resizebox{\linewidth}{!}{%
\begin{tabular}{@{}llll@{}}
\toprule
\textbf{Model} & \textbf{QASC} $\uparrow$  & \textbf{OBQA} $\uparrow$ & \textbf{ComQA} $\uparrow$  \\ \midrule
SMLM - Hypo  + CF          & 27.2 $\pm$ 0.6  & 34.6 $\pm$ 0.8 & 38.8 $\pm$ 0.6\\
SMLM - Quesn + CF          & 26.5 $\pm$ 1.2  & 32.2 $\pm$ 1.1 & 35.4 $\pm$ 1.3\\
SMLM - Hypo  + Rand Sample & 22.6 $\pm$ 1.4  & 28.4 $\pm$ 1.5 & 32.2 $\pm$ 1.4\\
SMLM - Gold F+ Hypo + CF   & 72.4 $\pm$ 0.8  & 75.2 $\pm$ 0.7 & - \\
\bottomrule
\end{tabular}%
}
\caption{Effect of Question to Hypothesis Conversion (Hypo), Curriculum Filtering (CF) and providing the Gold Fact context on the Validation split.}
\label{tab:abshypcf}
\end{table}

\paragraph{Effect of Sythetic Triple corpus size}
Figure \ref{fig:trainingcurve} compares our two modelling approaches when we train them with varying number of KTL training samples. NCE refers to our KRL model trained with NCELoss and Cosine similarity. We can observe our KRL model learns faster due to additional supervision, but the SMLM model performs the best, when trained with more samples. The performance tapers after $10^5$ samples, indicating the models are overfitting to the synthetic data. 

\paragraph{Error Analysis}
We sampled 50 error cases from each of our question answering tasks. Our KTL framework allows learning from knowledge graphs, that includes synthetic knowledge graphs. 
Both our instantiation, SMLM and KRL function as a knowledge base score generator, where given the inputs and a target the generator yields a score, how improbable is the target to be present in the knowledge base. 
Most of our errors are when all context, question and answer-option have a large distance scores, and the model accuracy degenerates to that of a random model. This larger distance indicates the model is highly perplexed to see the input text. 
For aNLI and SocialIQA, we possess relevant context and our performance is significantly better in these datasets, but for other tasks we have another source of error, i.e., context creation. In several cases the context is irrelevant and acts as noise. Other errors include, when the questions require complex reasoning such as understanding negation, conjunctions and disjunctions; temporal reasoning such ``6 am'' being before ``10 am'', and multi-hop reasoning. These complex reasoning tasks are required to answer a significant number of questions in the science and commonsense QA tasks.
We also tried to utilize a text generation model, such as GPT-2, to generate and compare with ground truth text using our KTL framework, but preliminary results show the model is overfitting to the synthetic dataset, and has a significantly low performance.

\paragraph{Other Instantiations} Our KTL framework can be implemented using other methods, such as using a Generator/Discriminator pre-training proposed in Electra~\cite{clark2019electra}, and  sequence-to-sequence methods. The distance functions for sequence-to-sequence models can be similar to our SMLM model, the cross-entropy loss for the expected generated sequence. Discriminator based methods can adapt the negative class probabilites as the distance function. Studying different instantiations and their implications are some of the interesting future works.

\section{Related Work}
\subsection{Unsupervised QA}
Recent work on unsupervised question answering approach the problem in two ways, a domain adaption or transfer learning problem \cite{chung-etal-2018-supervised}, or a data augmentation problem \cite{yang-etal-2017-semi,dhingra-etal-2018-simple,wang-etal-2018-multi-perspective,alberti-etal-2019-synthetic}. The work of \cite{lewis-etal-2019-unsupervised,fabbri2020template,puri2020training} use style transfer or template-based question, context and answer triple generation, and learn using these to perform unsupervised extractive question answering. There is also another approach of learning generative models, generating the answer given a question or clarifying explanations and/or questions, such as GPT-2 \cite{radford2019language} to perform unsupervised question answering \cite{shwartz2020unsupervised,bosselut2019dynamic,bosselut2019comet}. In contrast, our work focuses on learning from knowledge graphs and generate vector representations or sequences of tokens not restricted to the answer but including the context and the question using the masked language modeling objective.

\subsection{Use of External Knowledge for QA}
There are several approaches to add external knowledge into models to improve question answering. Broadly they can be classified into two, learning from unstructured knowledge and structured knowledge. In learning from unstructured knowledge, recent large pre-trained language models \cite{peters2018deep,radford2019language,devlin2018bert,liu2019roberta,clark2020electra,lan2019albert,m2020contextualized,bosselut2019comet} learn general-purpose text encoders from a huge text corpus. On the other hand, learning from structured knowledge includes learning from structured knowledge bases \cite{yang-mitchell-2017-leveraging,bauer-etal-2018-commonsense,mihaylov-frank-2018-knowledgeable,wang-jiang-2019-explicit,sun-etal-2019-pullnet} by learning knowledge enriched word embeddings. Using structured knowledge to refine pre-trained contextualized representations learned from unstructured knowledge is another approach \cite{peters-etal-2019-knowledge,yang-etal-2019-enhancing-pre,zhang-etal-2019-ernie,liu2019k}.

Another approach of using external knowledge includes retrieval of knowledge sentences from a text corpora  \cite{das2019multi,chen-etal-2017-reading,lee-etal-2019-latent,banerjee2019careful,banerjee2020knowledge,mitra2019exploring,banerjee2019asu}, or knowledge triples from knowledge bases \cite{min2019knowledge,wang2020k} that are useful to answer a specific question. Another recent approach is to use language model as knowledge bases \cite{petroni2019language}, where they query a language model to un-mask a token given an entity and a relation in a predefined template. In our work, we use knowledge graphs to learn a self-supervised generative task to perform zero-shot multiple-choice QA. 

\subsection{Knowledge Representation Learning}
Over the years there are several methods discovered to perform the task of knowledge representation learning.
Few of them are: TransE \cite{bordes2013translating} that views relations as a translation vector between head and tail entities, TransH \cite{wang2014knowledge} that overcomes TransE's inability to model complex relations, and TransD \cite{ji2015knowledge} that aims to reduce the parameters by proposing two different mapping matrices for head and tail.
KRL has been used in various ways to generate natural answers \cite{yin2016neural,he2017generating} and generate factoid questions \cite{serban2016generating}. The task of Knowledge Graph Completion~\cite{yao2019kg} is to either predict unseen relations  $r$  between two existing entities: $(h,?,t)$ or predict the tail entity $t$ given the head entity and the query relation: $(h,r,?)$. Whereas we are learning to predict including the head, $(?,r,t)$. In KTL, head and tail are not similar text phrases (context and answer) unlike Graph completion.  We further modify TransD and adapt it to our KTL framework to perform zero-shot QA.

\section{Conclusion}
In this work, we propose a new framework of Knowledge Triplet Learning over knowledge graph entities and relations. We show learning all three possible functions, $f_r$, $f_h$, and $f_t$ helps the model to perform zero-shot multiple-choice question answering, where we do not use question-answering annotations. We learn from both human-annotated and synthetic knowledge graphs and evaluate our framework on the six question answering datasets. Our framework achieves state-of-the-art in the zero-shot question answering task achieving performance like prior supervised work and sets a strong baseline in the few-shot question answering task. 

\section{Acknowledgements}
The authors acknowledge support from the DARPA SAIL-ON program, and ONR award N00014-20-1-2332. The authors will also like to thank Tejas Gokhale, Arindam Mitra and Sandipan Choudhuri for their feedback on earlier drafts. 

\bibliography{emnlp2020}
\bibliographystyle{acl_natbib}




\end{document}